# Factorization of Discrete Probability Distributions

**Dan Geiger**[*]
Computer Science Department
Technion
Haifa, 36000, Israel
dang@cs.technion.ac.il

**Christopher Meek**
Microsoft Research
Microsoft Cooperation
Redmond, WA 98052, USA
meek@microsoft.com

**Bernd Sturmfels**
Department of Mathematics
University of California Berkeley
Berkeley, CA 94720, USA
bernd@math.berkeley.edu

## Abstract

We formulate necessary and sufficient conditions for an arbitrary discrete probability distribution to factor according to an undirected graphical model, or a log-linear model, or other more general exponential models. This result generalizes the well known Hammersley-Clifford Theorem.

## 1 Introduction

In this paper we describe a class of exponential models for discrete distributions. These models include two important classes of models: log-linear models and undirected graphical models. We define these models in terms of a polynomial mapping from a set of parameters to distributions and analyze the algebraic properties of these models. Our analysis provides necessary and sufficient conditions for a discrete probability distribution to factor according to an undirected graphical model, or a log-linear model, or a more general exponential model. In particular, these conditions are shown to include constraints on some *cross product ratios*, in addition to independence statements. Our results generalize the Hammersley-Clifford Theorem which characterizes the factorization of strictly positive distributions with respect to undirected graphs (e.g., Besag 1974; Lauritzen 1996).

Graphical models have been defined and studied in the statistical literature in two distinct but related approaches. The first approach is to define undirected graphical models by specifying a graph according to which a probability distribution must factor in order to belong to the undirected graphical model. This direction was emphasized, for example, by Darroch, Lauritzen, and Speed (1980). The second approach is to define graphical models by specifying, through a graph, a set of conditional independence statements which a probability distribution must satisfy in order to belong to the graphical model. This direction was emphasized, for example, by Pearl (1988) and Geiger and Pearl (1993). Lauritzen (1996, Chapter 3) compared these approaches and herein we extend his results.

The paper is organized as follows. In Section 2, we define a class of exponential models and describe log-linear and undirected graphical models. In Section 3, we provide necessary and sufficient conditions for a discrete probability distribution to factor according to such an exponential model or to be the limit of distributions that factor. In Section 4, we focus our attention on undirected graphical models illustrating how the results of Section 3 generalize the Hammersley-Clifford Theorem for undirected graphical models. Finally, in Section 5 we present an open problem regarding undirected graphical models and provide some initial results towards its solution.

## 2 Technical Background

Our objects of study are certain statistical models for a finite state space $\mathcal{X}$. The class of models to be considered consists of discrete exponential families (models) of the form

$$P_\theta(x) = Z(\theta) \cdot e^{\langle \theta, T(x) \rangle}, \quad \theta \in [-\infty, \infty)^d \qquad (1)$$

where $x \in \mathcal{X}$, $Z(\theta)$ is a normalizing constant, $\langle \cdot, \cdot \rangle$ denotes an inner product and *sufficient statistics* $T : \mathcal{X} \mapsto \mathbf{N}^d \setminus \{\mathbf{0}\}$ where $\mathbf{N}$ denotes the set of non-negative integers and $\mathbf{0}$ is a vector of $d$ zeroes.

We find it convenient to reformulate these models as follows. We identify $\mathcal{X}$ with the set $\{1, 2, \ldots, m\}$. A probability distribution over $\mathcal{X}$ is then a vector $P = (p_1, \ldots, p_m)$ in $\mathbf{R}_{\geq 0}^m$ such that $p_1 + \cdots + p_m = 1$. The *support* of an $m$-dimensional vector $v$ is the set of indices $\mathrm{supp}(v) = \{i \in \{1, \ldots, m\} : v_i \neq 0\}$ and thus the *support* of a probability distribution $P$ is the set of indices $\mathrm{supp}(P) = \{i \in \{1, \ldots, m\} : p_i > 0\}$. Let $A = (a_{ij})$ be a non-negative integer $d \times m$-matrix. This

---

matrix defines the mapping $\phi_A : \mathbf{R}^d_{\geq 0} \mapsto \mathbf{R}^m_{\geq 0}$ which takes non-negative real $d$-vectors to non-negative real $m$-vectors via

$$(t_1, \ldots, t_d) \mapsto \left( \prod_{i=1}^d t_i^{a_{i1}}, \prod_{i=1}^d t_i^{a_{i2}}, \ldots, \prod_{i=1}^d t_i^{a_{im}} \right) \quad (2)$$

where $0^a = 0$ for all $a$. We say that a probability distribution $P$ *factors according to the model* $A$ (described by Equation 2) if and only if $P$ is in the image of the mapping $\phi_A$.

The class of models in Equation 1 and Equation 2 are identical. Note that each column of $A$ corresponds to a different state $x$ of $\mathcal{X}$. Thus, a model defined by Equation 1 with sufficient statistics $T(x)$ is equivalent to a model defined by Equation 2 with matrix $A$ if and only if the columns $a_j$ of $A$ coincide with the corresponding $T(x)$. For a particular distribution from a model of the form Equation 1 with sufficient statistics $T(x)$ and parameters $\theta$, the corresponding parameters $t_i$ for the corresponding model in Equation 2 are $t_i = \exp(\theta_i)$ where $\exp(-\infty) = 0$.

This class of models includes log-linear and undirected graphical models used in the analysis of multiway contingency tables. When analyzing multiway contingency tables, the state space is a product space $\mathcal{X} = \prod_{X_j \in \mathbf{X}} I_{X_j}$ where $\mathbf{X} = \{X_1, \ldots, X_n\}$ is a set of (random) variables and $I_{X_j}$ is the set of states for variable $X_j$. A log-linear model is defined by a collection $\mathcal{G} = \{\mathcal{G}_1, \ldots, \mathcal{G}_m\}$ of subsets of $\mathbf{X}$. We refer to the $\mathcal{G}_i$ as the *generators* of the log-linear model. A *log-linear model* for a set of generators $\mathcal{G}$ is defined as

$$P(x) \propto \prod_{\mathcal{G}_i \in \mathcal{G}} \psi_{\mathcal{G}_i}(x)$$

where $x \in \mathcal{X}$ is an instantiation of the variables in $\mathbf{X}$ and $\psi_{\mathcal{G}_i}(x)$ is a *potential function* that depends on $x$ only through the values of the variables in $\mathcal{G}_i$. This log-linear model can be represented in the following way by a matrix $A$ as in Equation 2. The columns of $A$ are indexed by $\mathcal{X} = \prod_{X_j \in \mathbf{X}} I_{X_j}$. The rows of $A$ are indexed by pairs consisting of a generator $\mathcal{G}_i$ and an element of $\prod_{X_j \in \mathcal{G}_i} I_{X_j}$. All entries of $A$ are either zero or one. The entry is one if and only if the second entry in the row label is equal to image of the column label under the projection from $\mathcal{X}$ to $\prod_{X_j \in \mathcal{G}_i} I_{X_j}$.

**Example 1** *The no-three-way interaction model for $X_1, X_2, X_3$, where each $X_i$ is a binary variable, has generators $\mathcal{G} = \{\{X_1, X_2\}, \{X_2, X_3\}, \{X_1, X_3\}\}$, and is represented by the matrix*

|  | $p_{000}$ | $p_{001}$ | $p_{010}$ | $p_{011}$ | $p_{100}$ | $p_{101}$ | $p_{110}$ | $p_{111}$ |
|---|---|---|---|---|---|---|---|---|
| $t_1 \equiv \psi_{\{1,2\}}(00)$ | 1 | 1 | 0 | 0 | 0 | 0 | 0 | 0 |
| $t_2 \equiv \psi_{\{1,2\}}(01)$ | 0 | 0 | 1 | 1 | 0 | 0 | 0 | 0 |
| $t_3 \equiv \psi_{\{1,2\}}(10)$ | 0 | 0 | 0 | 0 | 1 | 1 | 0 | 0 |
| $t_4 \equiv \psi_{\{1,2\}}(11)$ | 0 | 0 | 0 | 0 | 0 | 0 | 1 | 1 |
| $t_5 \equiv \psi_{\{2,3\}}(00)$ | 1 | 0 | 0 | 0 | 1 | 0 | 0 | 0 |
| $t_6 \equiv \psi_{\{2,3\}}(01)$ | 0 | 1 | 0 | 0 | 0 | 1 | 0 | 0 |
| $t_7 \equiv \psi_{\{2,3\}}(10)$ | 0 | 0 | 1 | 0 | 0 | 0 | 1 | 0 |
| $t_8 \equiv \psi_{\{2,3\}}(11)$ | 0 | 0 | 0 | 1 | 0 | 0 | 0 | 1 |
| $t_9 \equiv \psi_{\{1,3\}}(00)$ | 1 | 0 | 1 | 0 | 0 | 0 | 0 | 0 |
| $t_{10} \equiv \psi_{\{1,3\}}(01)$ | 0 | 1 | 0 | 1 | 0 | 0 | 0 | 0 |
| $t_{11} \equiv \psi_{\{1,3\}}(10)$ | 0 | 0 | 0 | 0 | 1 | 0 | 1 | 0 |
| $t_{12} \equiv \psi_{\{1,3\}}(11)$ | 0 | 0 | 0 | 0 | 0 | 1 | 0 | 1 |

*A probability distribution for three variables $P = (p_{000}, p_{001}, p_{010}, p_{011}, p_{100}, p_{101}, p_{110}, p_{111})$ factors in the no-three-way interaction model if and only if it lies in the image of the associated mapping $\phi_A : \mathbf{R}^{12}_{\geq 0} \mapsto \mathbf{R}^8_{\geq 0}$ defined by*

$$(t_1, \ldots, t_d) \mapsto (t_1 t_5 t_9, t_1 t_6 t_{10}, \ldots, t_4 t_8 t_{12}).$$

An important subclass of log-linear models are the undirected graphical models. Such a model is specified by an undirected graph $G$ with vertex set $\mathbf{X}$ and edge set $\mathbf{E}$. The *undirected graphical model* for the graph $G$ is the log-linear model in which the generators are the cliques (maximally connected subgraphs) of the undirected graph $G$. The matrix $A$ of Equation 2 is a function of the graph $G$ and we write it as $A(G)$. Example 1 shows a log-linear model that is not graphical.

**Example 2** *The three-variable-chain graphical model with graph $G$ having two edges $X_1 - X_2$ and $X_2 - X_3$ has generators $\mathcal{G} = \{\{X_1, X_2\}, \{X_2, X_3\}\}$. When each $X_i$ is a binary variable, the matrix $A(G)$ is identical to the first eight rows of the matrix of Example 1.*

We conclude this section by relating certain polynomials equations to conditional independence statements. Given three discrete variables $X, Y, Z$, we define

$$\text{cpd}(X = \{x, x'\}, Y = \{y, y'\} \mid Z = z) = \quad (3)$$
$$P(x, y, z) P(x', y', z) - P(x', y, z) P(x, y', z)$$

where $x$ and $x'$ are states of $X$ and $y$ and $y'$ are states of $Y$ and $z$ is a state of $Z$. We call these polynomials *cross-product differences (CPDs)*. Note that the cross-product differences are essentially the same as the *cross-product ratios*

$$\text{cpr}(X = \{x, x'\}, Y = \{y, y'\} \mid Z = z) = \quad (4)$$
$$\frac{P(x, y, z) P(x', y', z)}{P(x', y, z) P(x, y', z)}.$$

See e.g. (Lauritzen 1996, pp. 37). The notions of cpd and cpr are identical in the sense that

$$\text{cpd}(X = \{x, x'\}, Y = \{y, y'\} \mid Z = z) = 0$$

if and only if

$$\text{cpr}(X = \{x, x'\}, Y = \{y, y'\} | Z = z) = 1,$$

provided the denominators in (4) are nonzero. Sometimes it is more convenient to use cross-product ratios when interpreting higher degree binomials associated with an undirected graphical model. When $X$ and $Y$ each represent a single binary variable, we shorten the notation in (4) to

$$\text{cpr}(X, Y | Z = z) = \frac{P(x, y, z) P(x', y', z)}{P(x', y, z) P(x, y', z)}. \quad (5)$$

Let $X_1, \ldots, X_n$ denote discrete variables, where $I_{X_j}$ is the set of states of the variable $X_j$. We fix the polynomial ring $\mathbf{R}[\mathcal{X}]$ whose indeterminates $p_{a_1 a_2 \cdots a_n}$ are indexed by the joint states of $\mathcal{X} = I_{X_1} \times I_{X_2} \times \cdots \times I_{X_n}$. Conditional independence statements have the form

$$X \text{ is independent of } Y \text{ given } Z, \quad (6)$$

where $X$, $Y$ and $Z$ are pairwise disjoint subsets of $\{X_1, \ldots, X_n\}$. The statement (6) translates into a large set of CPDs of the form (3). Namely, we take $\text{cpd}(X = \{x, x'\}, Y = \{y, y'\} | Z = z)$, where $x, x'$ runs over distinct states in $\prod_{X_i \in X} I_{X_i}$, where $y, y'$ runs over distinct states in $\prod_{X_j \in Y} I_{X_j}$, and where $z$ runs over $\prod_{X_k \in Z} I_{X_k}$. The independence statement (6) is said to be *saturated* if $X \cup Y \cup Z = \{X_1, \ldots, X_n\}$. The CPDs associated with a saturated independence fact are all square-free quadratic binomial equations, namely, polynomials having exactly two monomials each consisting of two distinct terms (i.e., $t_1 t_2 - t_3 t_4 = 0$).

## 3 Distributions that Factor

In this section, we provide a characterization of those distributions that factor according to a model $A$ and of those distributions that are the limit of distributions that factor. These distributions lie in image($\phi_A$) where $\phi_A$ is the mapping defined by Equation 2.

We use basic notions of ideals, varieties, and ideal bases from computational algebraic geometry (e.g., Cox, Little, and O'Shea, 1997). We work in the ring $\mathbf{R}[x] = \mathbf{R}[x_1, \ldots, x_m]$ of polynomials with real coefficients in the indeterminates $x_1, \ldots, x_m$. An *ideal* $I$ is a subset of $\mathbf{R}[x]$ which satisfies three properties: (a) the zero polynomial is in $I$, (b) if $q_1, q_2 \in I$, then $q_1 + q_2 \in I$, and (c) if $b \in R[x]$, and $q \in I$, then $b \cdot q \in I$. With every ideal $I$ in $\mathbf{R}[x]$ we associate two *varieties*,

$$X^K = \{x \in K^m : q(x) = 0, \text{ for every } q \in I\}.$$

where $K$ denotes either the positive real numbers $\mathbf{R}_{>0}$ or the non-negative real numbers $\mathbf{R}_{\geq 0}$. Hence $X^{\geq 0}$ is the common zero set in $\mathbf{R}^m_{\geq 0}$ of all polynomials in $I$, and $X^{>0}$ is the common zero set in $\mathbf{R}^m_{>0}$ of all polynomials in $I$. Testing $x \in X^K$ is equivalent to checking that $q(x) = 0$ for all $x \in K^m$, for all $q \in I$. *Hilbert's Basis Theorem* states that every ideal in $\mathbf{R}[x]$ is *finally generated*, namely, every ideal $I$ in $\mathbf{R}[x]$ contains a finite subset $\{g_1, \ldots, g_n\}$, called an *ideal basis of $I$*, such that every $q \in I$ can be written as $q(x) = \sum_{i=1}^n b_i(x) g_i(x)$ where $b_i$ are polynomials in $\mathbf{R}[x]$. Consequently, a point $x$ in $K^m$ lies in $X^K$ if and only if $g_1(x) = \ldots = g_n(x) = 0$. The ideal generated by a set of polynomials $g = \{g_1, \ldots, g_n\}$ is denoted by $\langle g_1, \ldots, g_n \rangle$. An ideal $I$ is *prime* if whenever $q \cdot p \in I$, then either $q \in I$ or $q \in I$. We will focus on *toric ideals* which are prime ideals that have an ideal basis consisting of binomials of arbitrary degree.

Let $a_j = (a_{1j}, \ldots, a_{dj})$ denote the $j$-th column vector of the $d \times m$-matrix $A$. Note that $\text{supp}(a_j) \subseteq \{1, 2, \ldots, d\}$. A subset $F$ of $\{1, \ldots, m\}$ is said to be *nice* if, for every $j \in \{1, \ldots, m\} \setminus F$, the support $\text{supp}(a_j)$ of the vector $a_j$ is not contained in $\bigcup_{l \in F} \text{supp}(a_l)$.

**Lemma 1** *A probability distribution $P$ factors according to $A$ only if the support of $P$ is nice.*

*Proof:* Let $P$ be a probability distribution which factors according to $A$, that is, $P \in \text{image}(\phi_A)$. We must show that $F = \text{supp}(P)$ is nice. Let $(t_1, \ldots, t_d)$ be any preimage of $P$ under $\phi_A$. Then

$$p_j = \prod_{i=1}^d t_i^{a_{ij}} > 0 \quad \text{for } j \in F \quad (7)$$

and

$$p_j = \prod_{i=1}^d t_i^{a_{ij}} = 0 \quad \text{for } j \notin F. \quad (8)$$

Suppose that $F$ is not nice. Then $\text{supp}(a_k)$ is contained in $\bigcup_{l \in F} \text{supp}(a_l)$ for some $k \notin F$. Consequently for every $i \in \text{supp}(a_k)$, there exists an $f \in F$ such that $a_{if} > 0$. Hence, due to (7), $t_i > 0$ for every $i \in \text{supp}(a_k)$. Thus $p_k = \prod_{i \in \text{supp}(a_k)} t_i^{a_{ik}} > 0$ contrary to our assumption that $k \notin F$. //

The *non-negative toric variety* $X_A^{\geq 0}$ is the set of all vectors $(x_1, \ldots, x_m) \in \mathbf{R}^m_{\geq 0}$ which satisfy

$$x_1^{u_1} x_2^{u_2} \cdots x_m^{u_m} = x_1^{v_1} x_2^{v_2} \cdots x_m^{v_m} \quad (9)$$

whenever $u_1, \ldots, u_m, v_1, \ldots, v_m$ are non-negative integers and satisfy the linear relations

$$u_1 a_{i1} + u_2 a_{i2} + \cdots + u_m a_{im} = \quad (10)$$
$$v_1 a_{i1} + v_2 a_{i2} + \cdots + v_m a_{im}$$

for $i = 1, \ldots, d$. Note that (10) is equivalent in vector notation to

$$u_1 a_1 + u_2 a_2 + \cdots + u_m a_m = \\ v_1 a_1 + v_2 a_2 + \cdots + v_m a_m. \quad (11)$$

In this definition we adopt the convention $0^0 = 1$. Since the exponents $u_1, \ldots, u_m, v_1, \ldots, v_m$ used in (9) were assumed to be non-negative integers, the set $X_A^{\geq 0}$ is indeed an algebraic variety, that is, the zero set of a system of polynomial equations.

**Lemma 2** *A probability distribution $P$ factors according to $A$ only if $P$ lies in the non-negative toric variety $X_A^{\geq 0}$.*

*Proof:* We need to show that the image of $\phi_A$ is a subset of $X_A^{\geq 0}$. Indeed, suppose that $x = (x_1, \ldots, x_m) \in \text{image}(\phi_A)$. There exist non-negative reals $t_1, \ldots, t_d$ such that $x_i = t_1^{a_{1i}} t_2^{a_{2i}} \cdots t_d^{a_{di}}$ for $i = 1, \ldots, m$. This implies that (9) holds whenever (10) holds because

$$t_i^{u_1 a_{i1} + u_2 a_{i2} + \cdots + u_m a_{im}} = t_i^{v_1 a_{i1} + v_2 a_{i2} + \cdots + v_m a_{im}}$$

whenever (10) holds. Hence $x$ lies in $X_A^{\geq 0}$. //

The main contribution of this section is the formulation of necessary and sufficient conditions for a probability distribution to factor according to a matrix $A$. This result, when $A$ is appropriately selected, applies to undirected graphical models, log-linear models, and other statistical models.

**Theorem 3** *A probability distribution $P$ factors according to $A$ if and only if $P$ lies in the non-negative toric variety $X_A^{\geq 0}$ and the support of $P$ is nice.*

*Proof Outline:* The only-if direction has been proved in Lemmas 1 and 2. For the if-direction, fix any vector $P \in X_A^{\geq 0}$ whose support $F = \text{supp}(P)$ is nice. We claim that $P$ lies in the image of $\phi_A$, or equivalently that the system of equations (7, 8) has a non-negative real solution vector $(t_1, \ldots, t_d)$. This claim, along with other related results are proved in an extended version of this paper (Geiger, Meek, and Sturmfels, 2002).

We now turn our discussion to the set of distributions that are the limit of distributions that factor. In general, $\text{image}(\phi_A)$ is not a closed subset of the orthant $\mathbf{R}_{\geq 0}^m$. This is important because if there are distributions that do not factor according to a model but are the limit of distributions that do factor, then there are data sets for which the MLE does not exist.

The next theorem states that the set of probability distributions which lie in the toric variety $X_A^{\geq 0}$ coincide with those in the closure of the image of $\phi_A$—that is, $X_A^{\geq 0} = \text{closure}(\text{image}(\phi_A))$. This result means that $P \in X_A^{\geq 0}$ if and only if $P$ factors according to $A$, or $P$ is the limit of probability distributions which factor according to $A$. The set of distributions in $X_A^{\geq 0}$, when $A$ consists only of zeroes and ones, is called the *extended log-linear model* by Lauritzen (1996). Thus Theorem 4 below amounts to an algebraic description of extended exponential models and, thus, extended log-linear models and extended undirected graphical models.

**Theorem 4** *A probability distribution $P$ factors according to $A$ or is the limit of probability distributions that factor according to $A$ if and only if $P$ lies in the non-negative toric variety $X_A^{\geq 0}$.*

Theorems 3 and 4 together characterize probability distributions in $X_A^{\geq 0} \setminus \text{image}(\phi_A)$, namely, distributions that are the limit of factorizable distributions but do not factor themselves. These distributions are those that lie in $X_A^{\geq 0}$ but have a support which is not nice.

The task of checking that a point (e.g. a distribution) is in the zero set of all of the polynomials in an ideal, as required by Theorem 3 and Theorem 4, appears extremely hard, but there are two fundamental results which make it tractable. The first is *Hilbert's Basis Theorem* which states that every ideal in $\mathbf{R}[x]$ is finally generated. The second is Buchberger's algorithm that produces a distinguished ideal basis, called *Gröbner basis*, for any given ideal $I$. Variants of this algorithm have been implemented in virtually every symbolic algebra package available, albeit the implementations in MAPLE and MATHEMATICA are less efficient than those in some freely-available programs such as SINGULAR or COCOA. With this background, we now rephrase Theorem 3 and Theorem 4 as follows.

**Theorem 5** *A probability distribution $P$ factors according to an exponential model $A$ if and only if the support of $P$ is nice and all polynomials in an ideal basis of the toric ideal $I_A$ vanish at $P$.*

**Theorem 6** *A probability distribution $P$ is the limit of probability distributions that factor according to $A$ if and only if all polynomials in an ideal basis of the toric ideal $I_A$ vanish at $P$.*

We call these the *Factorization Theorem* and the *Limit Factorization Theorem* respectively. Thus, if we know a small ideal basis for $I_A$, which can be generated by a symbolic algebra program such as SINGULAR, then we can efficiently test whether or not a distribution $P$ lies in $X_A^{\geq 0}$ by checking that $P$ satisfies these polynomials.

It is important to note that one can often identify smaller sets than an ideal basis for $I_A$ for use in Theorems 5 and 6 when testing a distribution. In other words, we can identify a smaller set $B$ of polynomials such that $P \in X_A^{\geq 0}$ if and only if $P$ is the common zero set of the polynomial in $B$.

We will see that for undirected graphical models, the Hammersley-Clifford theorem, to be discussed in the next section, defines a small subset of binomials which do not generate the ideal $I_{A(G)}$, but their zero set does define $X_{A(G)}^{>0}$. Nevertheless, identifying an ideal basis for $I_A$, rather than a subset whose zero set defines the variety $X_A^{\geq 0}$ allows one to identify a complete set of moves for sampling from the conditional distribution of data given sufficient statistics for an exponential model, as described by Diaconis and Sturmfels (1998). This result complements alternative sampling methods (Besag and Clifford, 1989). We note that for directed graphical models, which are not discussed in this paper, direct sampling methods are well known (e.g., Lauritzen, 1996; Patefield, 1981).

## 4 The Hammersley-Clifford Theorem

The Hammersley-Clifford Theorem relates the factorization of a strictly positive distribution $P$ according to an undirected graphical model to a set of independence statements that must hold in $P$. In this section we describe the Hammersley-Clifford Theorem in the language of ideals and varieties and compare it to our Factorization theorem.

Let $G$ be an undirected graphical model with variables $\{X_1, \ldots, X_n\}$ as before. We define $I_{\text{pairwise}}(G)$ to be the ideal in $\mathbf{R}[\mathcal{X}]$ generated by the quadratic binomials corresponding to all the independence statements $X_i$ is independent of $X_j$ given $\{X_1, \ldots, X_n\} \setminus \{X_i, X_j\}$ where $(X_i, X_j)$ runs over all non-edges of the graph $G$. Note that this independence statement is saturated, so the polynomials are in fact binomials. The ideal $I_{pairwise(G)}$ defines the varieties $X_{\text{pairwise}(G)}^{\geq 0}$ and $X_{\text{pairwise}(G)}^{>0}$. Lauritzen (1996) uses the notation $M_P(\mathcal{G})$ to denote the variety $X_{\text{pairwise}}^{\geq 0}(G)$ and states the following three inclusions, which hold for every graph $G$, and are generally all strict:

$$\text{image}(\phi_{A(G)}) \subseteq X_{A(G)}^{\geq 0} \subseteq \tag{12}$$
$$X_{\text{global}(G)}^{\geq 0} \subseteq X_{\text{pairwise}(G)}^{\geq 0}.$$

The variety $X_{\text{global}(G)}^{\geq 0}$ corresponds to the ideal $I_{\text{global}}(G)$ generated by the quadratic binomials corresponding to all the independence statements (6) where $Z$ separates $X$ from $Y$ in the graph $G$. Probability distributions in $X_{\text{global}(G)}^{\geq 0}$ are said to satisfy the *global Markov property* (Lauritzen, 1996). The middle inequality is addressed by Matus and Studeny (1995).

**Example 3**
*The four-cycle undirected graphical model for binary variables with graph $G'$ having four edges $X_1 - X_2$, $X_2 - X_3$, $X_3 - X_4$ and $X_1 - X_4$ has generators $\mathcal{G} = \{\{X_1, X_2\}, \{X_2, X_3\}, \{X_3, X_4\}, \{X_1, X_4\}\}$. This graph has four maximal cliques, one for each edge. The probability distributions $P(x_1, x_2, x_3, x_4)$ defined by this model have the form*

$$\psi_{\{1,2\}}(x_1, x_2) \cdot \psi_{\{2,3\}}(x_2, x_3) \cdot$$
$$\psi_{\{3,4\}}(x_3, x_4) \cdot \psi_{\{1,4\}}(x_1, x_4).$$

*If all four variables are binary then the ideal $I_{\text{pairwise}(G')}$ we just defined equals*

$$\langle p_{1011}p_{1110} - p_{1010}p_{1111}, p_{0111}p_{1101} - p_{0101}p_{1111}, \\ p_{1001}p_{1100} - p_{1000}p_{1101}, p_{0110}p_{1100} - p_{0100}p_{1110}, \\ p_{0011}p_{1001} - p_{0001}p_{1011}, p_{0011}p_{0110} - p_{0010}p_{0111}, \\ p_{0001}p_{0100} - p_{0000}p_{0101}, p_{0010}p_{1000} - p_{0000}p_{1010}\rangle. \tag{13}$$

*This is a toric ideal in a polynomial ring in sixteen indeterminates:*

$$I_{\text{pairwise}(G')} \subset \mathbf{R}[p_{0000}, p_{0001}, p_{0010}, \ldots, p_{1111}].$$

*The left four binomials in (13) represent the statement "$X_2$ is independent of $X_4$ given $\{X_1, X_3\}$", and the right four binomials in (13) represent the statement "$X_1$ is independent of $X_3$ given $\{X_2, X_4\}$". The variety $X_{\text{pairwise}}^K(G)$ is the set of all points in $K^{16}$ which are common zeros of these eight binomials. Note that $X_{\text{pairwise}}^K(G) = X_{\text{global}}^K(G)$ for the four-cycle model and therefore, for this model, the right inclusion of Equation 12 is an equality.*

In what follows we shall see the crucial differences between $K = \mathbf{R}_{\geq 0}$ and $K = \mathbf{R}_{>0}$. The following theorem is well-known in the statistics literature; see e.g. (Lauritzen 1996, pp. 36).

**Theorem 7 (Hammersley-Clifford)** *Let $G$ be an undirected graphical model. A strictly positive probability distribution $P$ factors according to $A(G)$ if and only if $P$ is in the variety $X_{\text{pairwise}(G)}^{>0}$.*

This theorem can be rephrased as follows:

$$\text{image}^+(\phi_{A(G)}) = X_{A(G)}^{>0} = X_{\text{pairwise}(G)}^{>0}$$

where $\text{image}^+(\phi_{A(G)})$ is the set of strictly positive distributions in the image of $\phi_{A(G)}$.

Our Factorization Theorem 5 generalizes the Hammersley-Clifford Theorem in two respects. First, it does not require $P$ to be strictly positive. Second, it does not require the matrix $A$ to represent an undirected graphical model. The main advantage of the Hammersley-Clifford Theorem over the Factorization Theorem is computational. The set $I_{\text{pairwise}(G)}$ is easy to describe while one must usually resort to a symbolic algebra program to produce an ideal basis or a Gröbner basis for $I_A$.

The proof of the Hammersley-Clifford Theorem given in (Lauritzen 1996) actually establishes the following slightly stronger result: any integer vector in the kernel of the matrix $A(G)$ is an *integer* linear combination of the vectors $u - v$ corresponding to the binomials $p^u - p^v$ arising from the conditional independence statements for the non-adjacent pairs $(X_i, X_j)$ in $G$. Translating this statement from the additive notation into multiplicative notation, we obtain the following:

**Corollary 8** *A binomial $p^u - p^v$ lies in the toric ideal $I_{A(G)}$ of an undirected graphical model $A(G)$ if and only if some monomial multiple of it, i.e., a binomial of the form $p^{u+w} - p^{v+w}$, lies in $I_{\text{pairwise}}(G)$.*

This corollary is important for computational purposes because it means that we can use the quadratic binomials in $I_{\text{pairwise}}(G)$ as input when computing the toric ideal $I_{A(G)}$, as outlined in the explanation of Proposition 9 below.

## 5 An open problem

We now discuss undirected graphical models from the perspective of the Factorization Theorem 5. We show that there are polynomials which do not correspond to independence statements that a probability distribution must satisfy in order to factor according to an undirected graphical model. This stands in sharp contrast to the Hammersley-Clifford Theorem which shows that a strictly-positive probability distribution must merely satisfy the pairwise independence statements in order to factor according to an undirected graphical model.

We first focus on the four-cycle model. Probability distributions which factor according to the four-cycle model of Example 3 must satisfy not just the eight quadratic binomials in (13), which arise from pairwise independence statements, but they must satisfy certain additional polynomials of degree 4 (namely, quartics) listed in Equation 14.

**Proposition 9** *Consider the four-cycle undirected graphical model of Example 3 with graph $G'$. A probability distribution $P$ factors according to the four-cycle or is the limit of probability distributions that factor according to the four-cycle if and only if $P$ satisfies the following basis of the ideal $I_{A(G')}$ consisting of $I_{\text{pairwise}(G')}$ along with*

$$\langle f_{12}^{\text{diff}}, f_{23}^{\text{diff}}, f_{34}^{\text{diff}}, f_{14}^{\text{diff}}, f_{12}^{\text{same}}, f_{23}^{\text{same}}, f_{34}^{\text{same}}, f_{14}^{\text{same}} \rangle$$

*where*

$$\begin{aligned}
f_{12}^{\text{diff}} &= p_{0100}p_{0111}p_{1001}p_{1010} - p_{0101}p_{0110}p_{1000}p_{1011}, \\
f_{23}^{\text{diff}} &= p_{0010}p_{0101}p_{1011}p_{1100} - p_{0011}p_{0100}p_{1010}p_{1101}, \\
f_{34}^{\text{diff}} &= p_{0001}p_{0110}p_{1010}p_{1101} - p_{0010}p_{0101}p_{1001}p_{1110}, \\
f_{14}^{\text{diff}} &= p_{0001}p_{0111}p_{1010}p_{1100} - p_{0011}p_{0101}p_{1000}p_{1110}, \\
f_{12}^{\text{same}} &= p_{0000}p_{0011}p_{1101}p_{1110} - p_{0001}p_{0010}p_{1100}p_{1111}, \\
f_{23}^{\text{same}} &= p_{0000}p_{0111}p_{1001}p_{1110} - p_{0001}p_{0110}p_{1000}p_{1111}, \\
f_{34}^{\text{same}} &= p_{0000}p_{0111}p_{1011}p_{1100} - p_{0011}p_{0100}p_{1000}p_{1111}, \\
f_{14}^{\text{same}} &= p_{0000}p_{0110}p_{1011}p_{1101} - p_{0010}p_{0100}p_{1001}p_{1111}.
\end{aligned}$$
(14)

This proposition is proved by an explicit machine calculation, by inputing to Algorithm 12.3 of (Sturmfels 1996) the eight quadratic generators of $I_{\text{pairwise}(G')}$. We did this using the symbolic algebra package called COCOA. The use of Algorithm 12.3 with the eight quadratic generators of $I_{\text{pairwise}(G')}$ as input is much more efficient than directly computing an ideal basis for the toric ideal $I_{A(G')}$.

Next we provide an interpretation of the ideal basis of the four-cycle given in Proposition 9. The basis prescribed by (14) has a meaningful statistical interpretation using cross product ratios. Adopting the definition of cpr in (5), the eight new basis elements (14) can be rewritten as follows:

$$\begin{aligned}
\text{cpr}(X_3, X_4 | X_1 X_2 = 01) / \text{cpr}(X_3, X_4 | X_1 X_2 = 10) &= 1, \\
\text{cpr}(X_1, X_4 | X_2 X_3 = 01) / \text{cpr}(X_1, X_4 | X_2 X_3 = 10) &= 1, \\
\text{cpr}(X_1, X_2 | X_3 X_4 = 01) / \text{cpr}(X_1, X_2 | X_3 X_4 = 10) &= 1, \\
\text{cpr}(X_2, X_3 | X_1 X_4 = 01) / \text{cpr}(X_2, X_3 | X_1 X_4 = 10) &= 1, \\
\text{cpr}(X_3, X_4 | X_1 X_2 = 00) / \text{cpr}(X_3, X_4 | X_1 X_2 = 11) &= 1, \\
\text{cpr}(X_1, X_4 | X_2 X_3 = 00) / \text{cpr}(X_1, X_4 | X_2 X_3 = 11) &= 1, \\
\text{cpr}(X_1, X_2 | X_3 X_4 = 00) / \text{cpr}(X_1, X_2 | X_3 X_4 = 11) &= 1, \\
\text{cpr}(X_2, X_3 | X_1 X_4 = 00) / \text{cpr}(X_2, X_3 | X_1 X_4 = 11) &= 1.
\end{aligned}$$
(15)

Note that (14) is obtained from (15) by multiplying the equations by the relevant denominators and that for (15) to be always defined, special care must be taken in defining division by zero.

Proposition 9 provides an ideal basis for the four-cycle undirected graphical model (with binary variables), however, the problem of explicitly providing a basis for an arbitrary undirected graphical model remains open. See (Takken, 1999) for related computations, and the work in (Sullivant and Hosten, 2002).

**Open problem:** Explicitly specify an ideal basis for the toric ideal $I_{A(G)}$ where $G$ is an arbitrary undirected graphical model for variables $X_1, \ldots, X_n$.

We examined quite a few examples of undirected graphical models and computed their ideal bases. In all these examples, we discovered the ideal basis elements have the form of ratio of ratio (recursively) of CPRs. For example, consider the four cycle $X_1 - X_2 - X_3 - X_4 - X_1$ in which $X_1$ and $X_2$ are binary but $X_3$ and $X_4$ have three states $\{0, 1, 2\}$. Here the ideal basis consists of 36 binomials of degree 2, representing independence statements of the form cpr $= 1$, 252 bi-nomials of degree 4, representing ratios of CPRs of the form (cpr/cpr) = 1 (as in (15)), and 12 binomials of degree 6 of the form (cpr/cpr)/(cpr/cpr) = 1, such as the following binomial equation,

$$p_{0102}p_{0110}p_{0121}p_{1001}p_{1012}p_{1020} -$$
$$p_{0101}p_{0112}p_{0120}p_{1002}p_{1010}p_{1021} = 0$$

which can be written as follows

$$\text{cpr}(X_3 = \{0,2\}; X_4 = \{2,1\}|X_1X_2 = 01)/$$
$$\text{cpr}(X_3 = \{2,1\}; X_4 = \{0,2\}|X_1X_2 = 01)$$

divided by

$$\text{cpr}(X_3 = \{0,2\}; X_4 = \{2,1\}|X_1X_2 = 10)/$$
$$\text{cpr}(X_3 = \{2,1\}; X_4 = \{0,2\}|X_1X_2 = 10))$$

equals 1.

We note that there is no general upper bound for the degrees of the binomials in the ideal basis of an undirected graphical model. For instance, if each variable in the four-cycle model has $p$ states, then there exists a minimal generator of degree $\geq p$. Such a binomial can be derived from Proposition 14.14 in (Sturmfels, 1996). It is interesting to note that the degrees in the ideal basis are also unbounded when the complexity of the model increases but all variables remain binary.

**Proposition 10** *There exists a undirected graphical model for $2n$ binary variables $X_1, \ldots, X_{2n}$ whose ideal basis contains a binomial of degree $2^n$.*

*Proof:* Let $G$ be the undirected graphical model whose only non-edges are $\{X_i, X_{i+n}\}$ for $i = 1, 2, \ldots, n$. Thus this model represents $n$ pairs of non-interacting binary variables. Let $p^u$ denote the product of all indeterminates $p_{i_1 \ldots i_{2n}}$ such that $i_1 = i_3 = i_5 = \cdots = i_{2n-1}$ and $i_1$ has the same parity as $i_2 + i_4 + i_6 + \cdots + i_{2n}$, and let $p^v$ denote the product of all indeterminates $p_{i_1 \ldots i_{2n}}$ such that $i_1 = i_3 = i_5 = \cdots = i_{2n-1}$ and $i_1$ has parity different from $i_2 + i_4 + i_6 + \cdots + i_{2n}$. Then $p^u - p^v$ is a binomial of degree $2^n$ which lies in the toric ideal $I_{A(G)}$. It can be checked, for instance using Corollary 12.13 in (Sturmfels, 1996), that $p^u - p^v$ is a minimal generator of $I_{A(G)}$. //

The undirected graphical models in the previous proof provide an interesting family for further study. Note that for $n = 2$ this is precisely the four-cycle model, and for $n = 3$ this is the edge graph of the octahedron, with cliques $\{1,3,5\}, \{1,3,6\}, \{1,4,5\}, \{1,4,6\}, \{2,3,5\}, \{2,3,6\}, \{2,4,5\}, \{2,4,6\}$. Here the binomial $p^u - p^v$ constructed in the proof of Proposition 10 equals

$$p_{000000} \cdot p_{000101} \cdot p_{010001} \cdot p_{010100} \cdot$$
$$\cdot p_{101011} \cdot p_{101110} \cdot p_{111010} \cdot p_{111111} -$$
$$p_{000001} \cdot p_{000100} \cdot p_{010000} \cdot p_{010101}$$
$$\cdot p_{101010} \cdot p_{101111} \cdot p_{111011} \cdot p_{111110}$$

which can also be written as ratio of ratios of CPRs. We note that providing an ideal basis even for this "$n$ non-interacting pairs" model is an open problem.

**Remark.** It has been brought to our attention by the reviewers that the work by Ripley and Kelly (1977) on heredity subsets and of Barndhoff-Nielsen (1978) on weak closures of exponential families are possibly related to the results presented herein. At time of publication, we have not had the chance to firm these relationships.